\documentclass[conference]{IEEEtran}
\usepackage{cite}
\usepackage{amsmath,amssymb,amsfonts}
\usepackage{algorithmic}
\usepackage{graphicx}
\usepackage{microtype}

\usepackage{textcomp}
\usepackage{xcolor}
\usepackage{booktabs}
\usepackage{array}
\usepackage{tabularx}
\usepackage{tikz}
\usepackage{url}
\usepackage[table]{xcolor}
\usetikzlibrary{arrows.meta,positioning,fit}

\definecolor{bestblue}{HTML}{0057B8}
\definecolor{secondblue}{HTML}{2F80ED}

\newcommand{\best}[1]{\textcolor{bestblue}{\bfseries #1}}
\newcommand{\second}[1]{\textcolor{secondblue}{#1}}

\newcommand{\state}{\mathbf{z}}

\def\BibTeX{{\rm B\kern-.05em{\sc i\kern-.025em b}\kern-.08em
    T\kern-.1667em\lower.7ex\hbox{E}\kern-.125emX}}

\begin{document}

\title{Visual Representation and History Modeling\\for Navigation World Models}

\author{
\IEEEauthorblockN{
\begin{minipage}{0.32\linewidth}
\centering
1\textsuperscript{st} Guangfu Guo\\
\textit{Clemson University}\\
Clemson, SC, USA\\
gguo@clemson.edu
\end{minipage}
\hfill
\begin{minipage}{0.32\linewidth}
\centering
2\textsuperscript{nd} Xiaoqian Lu\\
\textit{University of Ottawa}\\
Ottawa, ON, Canada\\
xlu053@uottawa.ca
\end{minipage}
\hfill
\begin{minipage}{0.32\linewidth}
\centering
3\textsuperscript{rd} Rui Liu\\
\textit{Clemson University}\\
Clemson, SC, USA\\
rliu3@clemson.edu
\end{minipage}
}

\vspace{1em}

\IEEEauthorblockN{
\begin{minipage}{0.32\linewidth}
\centering
4\textsuperscript{th} Yutong Chen\\
\textit{Clemson University}\\
Clemson, SC, USA\\
yutong2@clemson.edu
\end{minipage}
\hfill
\begin{minipage}{0.32\linewidth}
\centering
5\textsuperscript{th} Kunpeng Liu\\
\textit{Clemson University}\\
Clemson, SC, USA\\
kunpenl@clemson.edu
\end{minipage}
\hfill
\begin{minipage}{0.32\linewidth}
\centering
6\textsuperscript{th} Long Cheng\\
\textit{Clemson University}\\
Clemson, SC, USA\\
lcheng2@clemson.edu
\end{minipage}
}
}


\maketitle

\begin{abstract}
Navigation World Models (NWMs) predict action-conditioned visual futures for planning. Two practical challenges are central to their design: selecting a suitable visual representation and efficiently modeling observation history for repeated candidate queries. Standard Global-Softmax attention provides flexible interactions but repeatedly processes the same history, leading to increasing computation and memory costs for long contexts and multi-query planning. We study both problems within a unified conditional flow-transformer framework. We first compare five frozen visual representations under the same dynamics model and evaluation. To reduce redundant history computation, we design Cached-Linear, a hybrid architecture that combines local and shifted-window attention for target mixing with linear attention for reusable history access. We further develop Balanced Gated Delta Network (GDN), which augments this design with frame-wise recurrent memory for temporal history modeling. Experiments on RECON, SACSoN, and SCAND show that representation choice depends on the prediction objective: PAE-L performs best for reconstruction, RAE-B for direct prediction, and V-JEPA for long-horizon rollout. Under shared-history workloads, Cached-Linear substantially reduces computation and memory compared with Global-Softmax, while Balanced GDN improves selected direct-prediction endpoints with efficient context reuse. Overall, we systematically study visual representation and history modeling for NWMs and develop hybrid reusable-history architectures for efficient long-context and multi-query prediction.

\end{abstract}

\begin{IEEEkeywords}
navigation world models, visual representation, autoregressive
rollout, linear attention
\end{IEEEkeywords}

\section{Introduction}
\label{sec:introduction}

Navigation World Models (NWMs) predict future visual observations conditioned
on actions and use these predictions for forecasting and planning
~\cite{bar2025_nwm}. Instead of predicting directly in pixel space, recent
methods increasingly use features from pretrained visual encoders as the model
state~\cite{zhou2024dino,zhang2026raenwm}. This raises an important question:
what kind of visual representation is most suitable for predicting future
observations? A good representation for reconstruction is not necessarily a good
representation for prediction. Reconstruction mainly measures how much visual
information is preserved by the encoder and decoder. In contrast, a world
model must learn how the representation changes under different actions.
This becomes even harder in autoregressive rollout, where previous predictions
are added back into the history and used to generate later predictions.
Small errors can therefore accumulate over time. As a result, reconstruction,
direct prediction, and long-horizon rollout may favor different visual
representations. Selecting a representation only by reconstruction quality can
therefore be misleading for navigation world models.

Scaling NWMs to long observation histories introduces a second challenge: the same history is often reused across many candidate actions during planning, leading to substantial redundant computation. Standard global attention can model target–history interactions effectively, but it recomputes much of the same history-dependent information for every candidate action. This cost grows rapidly with both history length and the number of candidate queries. A useful history mechanism should therefore preserve prediction quality while allowing shared history computation to be reused efficiently. 

\begin{figure*}[!t]
    \centering
    \includegraphics[width=0.95\textwidth]{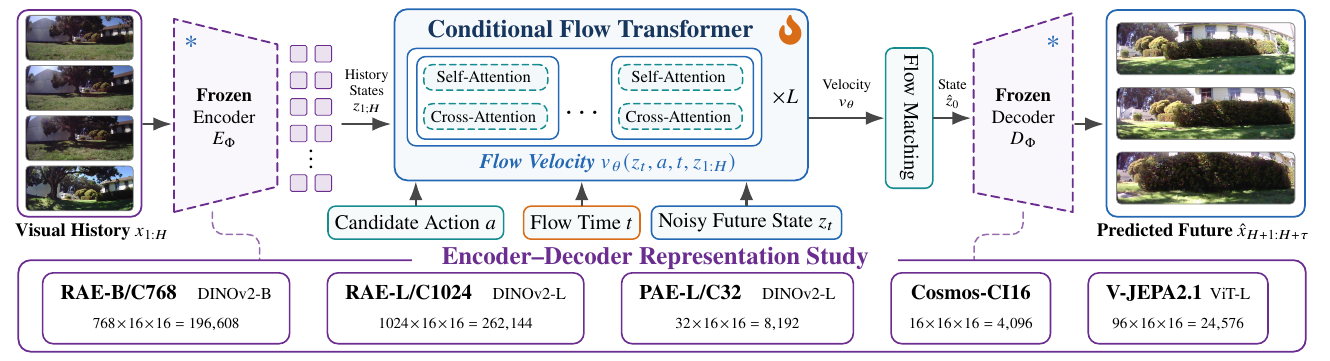}

\caption{\textbf{Overview of the Navigation World-model Framework and Visual-Representation Comparison.}
Visual history is encoded by a frozen encoder into latent states, and a conditional flow transformer predicts the future state using the observation history, candidate action, prediction horizon, and flow time. A matched frozen decoder maps the predicted latent back to image space. The same dynamics model and evaluation pipeline are used across five frozen visual representation---RAE-B, RAE-L, PAE-L, Cosmos, and V-JEPA---which differ in representation type and latent size.}

    \label{fig:nwm_figure_1}
    \vspace{-10pt}
\end{figure*}

In this work, we study visual representation and history modeling within the same conditional flow-transformer framework. We first benchmark five frozen visual representations using the same dynamics model, training setup, and decoded-image evaluation. These representations span dense semantic features, compact latent representations, and features learned through video prediction, allowing us to examine how different predictive state spaces affect reconstruction, direct prediction, and autoregressive rollout. To reduce redundant history computation during repeated-query planning, we adopt linear attention for reusable history access. Unlike full Softmax attention, linear attention compresses the observation history into a compact summary that can be computed once and reused across candidate queries. We further incorporate the Gated Delta Network (GDN), a recurrent mechanism that maintains a compact state through gated updates and accumulates temporal information frame by frame. Based on these mechanisms, we design two history-processing architectures. \emph{Cached-Linear} combines reusable linear history attention with local and shifted-window target mixing, while \emph{Balanced GDN} further integrates frame-wise GDN memory with local, shifted-window, linear, and global attention. We compare both designs against a standard \emph{Global-Softmax} baseline and evaluate their trade-offs in prediction quality and repeated-query efficiency.

Our experiments on RECON, SACSoN, and SCAND show that visual representation and history processing affect different aspects of navigation world-model performance. Different visual representations perform best for reconstruction, direct prediction, and long-horizon rollout, confirming that reconstruction quality alone does not determine predictive performance. For history processing, Global-Softmax remains competitive for a single short query, while Cached-Linear becomes substantially more efficient when many candidate actions share the same observation history. Balanced GDN introduces recurrent temporal memory and improves some direct-prediction results, particularly RAE-B Direct@32. Overall, these results characterize the effects of visual representation and history modeling on prediction quality, rollout behavior, and repeated-query efficiency.

Our main contributions are:

\begingroup
\setlength{\topsep}{2pt}
\setlength{\itemsep}{2pt}
\setlength{\parsep}{0pt}
\setlength{\partopsep}{0pt}

\begin{itemize}
\item We provide a controlled benchmark of five frozen visual 
representations within the same navigation world-model framework, evaluating 
reconstruction, direct prediction, and autoregressive rollout under a unified 
dynamics model, training protocol, and decoded-image evaluation.

\item We design two reusable history-processing mechanisms for long-history 
and repeated-query planning. \emph{Cached-Linear} amortizes history computation 
across candidate queries through a reusable linear summary, while 
\emph{Balanced GDN} introduces frame-wise recurrent memory for temporal 
history modeling.

\item Experiments on RECON, SACSoN, and SCAND characterize how visual 
representation and history modeling affect complementary aspects of navigation 
world-model performance, including prediction quality, rollout stability, and 
repeated-query efficiency.
\end{itemize}
\endgroup

\section{Related Work}
\label{sec:related}

\subsection{Navigation World Models}
NWM introduced action-conditioned visual generation for
planning~\cite{bar2025_nwm}. DINO-WM instead predicts in self-supervised
feature space and demonstrated zero-shot planning~\cite{zhou2024dino}.
RAE-NWM combines frozen DINOv2 features~\cite{oquab2023_dinov2} with a frozen
pixel decoder and flow-matched conditional diffusion-transformer dynamics
~\cite{zhang2026raenwm}. These methods motivate semantic state spaces, but
visual-state size or reconstruction quality alone need not determine
dynamics modelability. Recent navigation world models address rollout drift through model-generated
histories, sparse anchors, and geometry-aware guidance, while related systems
reduce planning cost through learned or calibrated objectives~\cite{yang2026arforcing,luan2026driftresistant,chahe2026monotone}. Other
work models uncertainty or predicts world and action jointly~\cite{zhu2026uanwm,zhou2026uninav,azuma2026navwam}. These
approaches target closed-loop navigation utility. We isolate decoded prediction quality, condition sensitivity, and
reusable-history execution. 

\subsection{Visual-Representation}
Representation autoencoders replace variational latents with pretrained
semantic encoders and learned decoders~\cite{zheng2025rae, lu2026ssr}, while PAE aligns a
compact latent with a representation prior for diffusion-friendly geometry
~\cite{yue2026pae}. Latent diffusion~\cite{rombach2022high}, DiT
~\cite{peebles2023scalable}, and the Diffusion Decoder Transformer
~\cite{wang2025ddt} likewise show that tokenizer and denoiser architecture
interact. REPA aligns intermediate diffusion features to frozen visual
features, and iREPA adds spatial projection and normalization
~\cite{yu2024repa,singh2025irepa}. A controlled robotic latent-diffusion study distinguishes visual fidelity,
downstream utility, and visual-representation quality
~\cite{nilaksh2026reconstruction}.

\subsection{Efficient Attention and History Modeling}

Standard Transformer attention provides flexible global interaction but has
quadratic cost with sequence length~\cite{vaswani2017attention}. Kernelized
linear Transformers reduce this cost by reorganizing attention computation and
avoiding the explicit construction of the full query--key matrix
~\cite{katharopoulos2020transformers}. Local-attention methods provide another
efficient alternative: Swin Transformer restricts self-attention to local
windows and uses shifted windows across layers to enable cross-window
communication~\cite{liu2021swin}.

Recurrent models offer a complementary approach to long-context processing.
Gated DeltaNet combines gated memory control with delta-rule updates to maintain
a compact recurrent state~\cite{yang2025gated}. SANA-WM applies frame-wise
Gated DeltaNet together with periodic Softmax attention for efficient
long-context world modeling~\cite{zhu2026sanawm}, while DeltaFlow extends
Gated Delta Networks to bidirectional continuous-flow denoising
~\cite{guo2026deltaflow}. Together, these works motivate hybrid designs that
combine global attention, local attention, linear attention, and recurrent
memory. However, how to design efficient navigation world models remains underexplored.

\section{Method}
\label{sec:method}

\begin{figure*}[!t]
    \centering
    \includegraphics[width=0.9\textwidth]{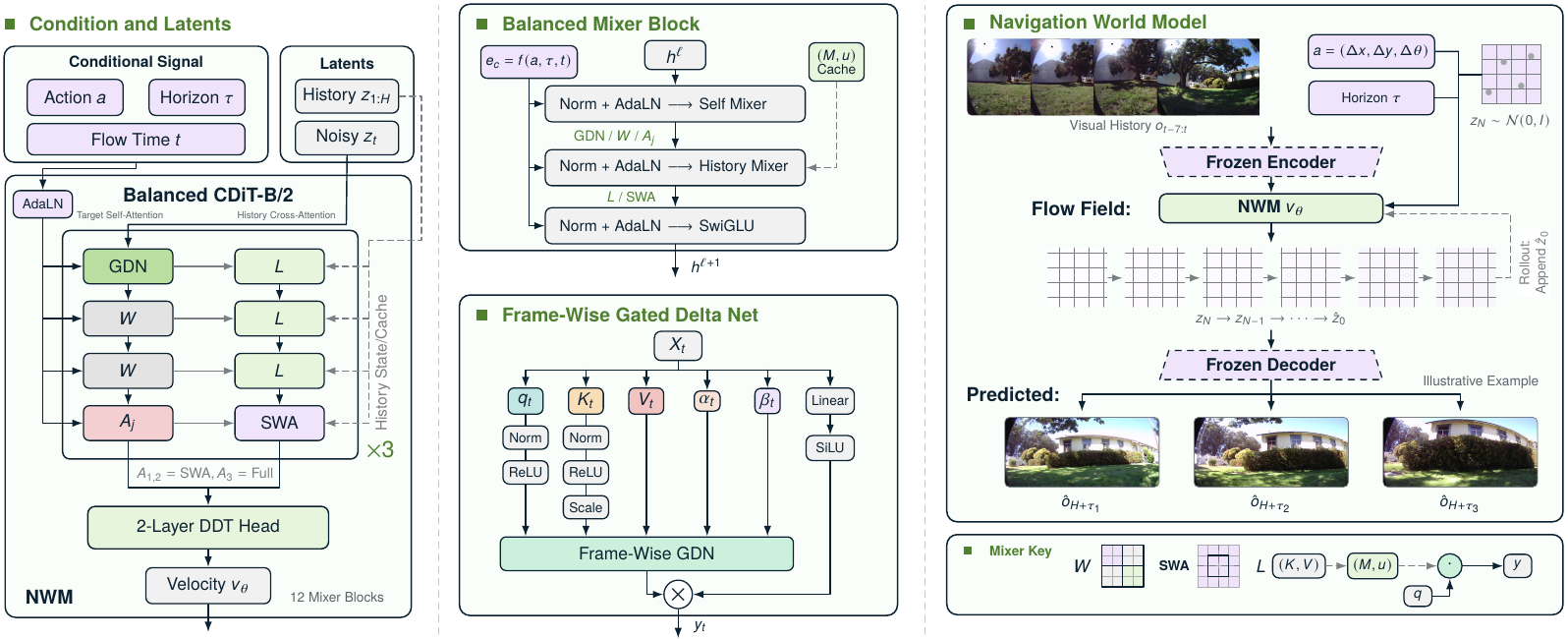}
    \vspace{-5pt}
\caption{Balanced GDN Dynamics Architecture. Left: action-conditioned
conditional flow prediction with frozen visual history latents. Center:
each CDiT block separates target mixing, reusable history mixing, and
feed-forward computation with frame-wise GDN memory updates. Right:
rollout updates the latent history directly. The key denotes window
attention ($W$), shifted-window attention (SWA), cached-linear summary
$(M,u)$, and GDN state $(S,r)$.}
\vspace{-10pt}
    \label{fig:gdn-architecture}
    
\end{figure*}

\subsection{Navigation World Model}
Given $H$ recent camera observations, a relative planar action, and a
requested future time, the model predicts the observation at that future time.
A frozen visual encoder first converts every image into a spatial latent grid.
We train the dynamics model by conditional flow matching. It learns to move
a noisy future latent back toward the clean latent while conditioning on the
observation history, action, prediction horizon, and flow time. Let
$c=(E(x_{1:H}),a,\tau,t)$ collect these conditions. With clean future latent
$\state_0$, Gaussian noise $\epsilon$, and interpolation time $t$, the flow
path and its target velocity are
\begin{equation}
\state_t=(1-t)\state_0+t\epsilon,\qquad \mathbf{v}^{\star}=\epsilon-\state_0.
\label{eq:flow-path}
\end{equation}
The model regresses this velocity with
\begin{equation}
\mathcal{L}_{\mathrm{FM}}=\mathbb{E}\left\|v_\theta(\state_t,c)-\mathbf{v}^{\star}\right\|_2^2.
\label{eq:fm-loss}
\end{equation}
At inference, Euler integration starts from $\hat{\state}_N\sim\mathcal N(0,I)$
and follows the learned field from $t=1$ to $t=0$:
\begin{equation}
\hat{\state}_{n-1}=\hat{\state}_{n}-\Delta t\,v_\theta(\hat{\state}_{n},c_n).
\label{eq:euler-sampling}
\end{equation}
A two-layer DDT head produces each velocity estimate, and the frozen decoder
maps the final latent back to an image.

The dynamics backbone is a 12-block CDiT-B/2 with width 768, 12 attention
heads, and a $16\times16$ token grid. A shallow, wider DDT prediction head has
two blocks of width 2,048 with 16 heads. History latents enter a separate
context stream, while the noisy future latent supplies the target stream.
The action, requested horizon, and flow time modulate the target blocks; the
history computation is condition-independent. This separation is important
for repeated queries because the same encoded observation window can serve
several horizons, integration steps, or candidate actions.

We use two inference modes. \emph{Direct prediction} generates each requested
horizon independently from real observations. \emph{Autoregressive rollout}
predicts a short step, inserts the predicted latent into the history, and
repeats. The latter does not decode and re-encode intermediate predictions.
Direct prediction measures endpoint accuracy from clean context. Rollout also
tests stability after model-generated states enter later queries.

\subsection{Frozen Visual Representation}
All visual representation produce a $16\times16$ spatial grid and remain frozen
during dynamics training. They differ in pretraining objective, channel count,
and decoder design. Table~\ref{tab:representation-interfaces} identifies each
frozen checkpoint and reports its reconstruction-reference DINO.

The dense representation retain native DINOv2 patch
states~\cite{oquab2023_dinov2,zheng2025rae}. The compact-aligned representation
uses the same semantic encoder family as the larger dense model but learns a
32-channel bottleneck intended for diffusion~\cite{yue2026pae}. The
generative-compact representation is the released continuous Cosmos image
tokenizer~\cite{nvidia2025cosmostokenizer}. The predictive-video representation
combines V-JEPA~2.1 features~\cite{murlabadia2026vjepa21} with an S-VAE
adapter and pixel decoder following~\cite{nilaksh2026reconstruction,zhang2025svae}.
We use these representation to compare
semantic, generative, and temporally predictive pretraining under the same
dynamics budget.

Raw latent errors lack a common scale because the representation differ in width,
normalization, and geometry. We instead report two image-space
quantities. The \emph{reconstruction reference} decodes the encoder output of
the real future image and measures information retained by the frozen
representation. The \emph{end-to-end prediction} decodes the dynamics model output
and combines representation, dynamics, and decoder error:
\begin{equation}
\resizebox{0.98\columnwidth}{!}{$\displaystyle
x_{\mathrm{rec}}=D(E(x)),\quad \hat{x}=D(\hat{\state}_0),\quad
m_{\mathrm{rec}}=m(x_{\mathrm{rec}},x),\quad m_{\mathrm{e2e}}=m(\hat{x},x)$}.
\label{eq:decoded-metrics}
\end{equation}
The reconstruction reference is useful context, not a mathematical lower
bound on the end-to-end score.

\subsection{Action Conditioning}
An action is the relative forward, lateral, and heading displacement
$a=(\Delta x,\Delta y,\Delta\theta)$. Each scalar passes through fixed Gaussian
Fourier features and a learned multilayer perceptron; the three outputs are
concatenated, fused with horizon and flow-time embeddings, and injected into
the target branch through adaptive transformer conditioning. The history
branch depends only on the observation window, so its computation can be
reused across candidate actions.

\subsection{History Mixer Architectures}

Global-Softmax, which applies full attention between the target and history
streams. It is effective for short, single-query
prediction, but repeatedly recomputes interactions with the same observation
history across flow-integration steps and candidate actions. To reduce this
redundancy and support longer observation histories, we design two reusable
history-processing architectures: \emph{Cached-Linear} and
\emph{Balanced GDN}.
Table~\ref{tab:mixer-components} summarizes the components used by these
architectures.

\begin{table}[t]
\centering
\caption{History-mixing components and their roles.}
\label{tab:mixer-components}
\footnotesize
\renewcommand{\arraystretch}{1.15}
\setlength{\tabcolsep}{4pt}
\begin{tabularx}{\columnwidth}{
    @{}
    >{\raggedright\arraybackslash\bfseries}p{0.32\columnwidth}
    >{\raggedright\arraybackslash}X
    @{}
}

\rowcolor{white}
\toprule
Component & Function \\
\midrule

\rowcolor{gray!10}
Window attention ($W$)
& Exact self-attention within $8\times8$ windows for fine-grained local mixing. \\

Shifted-window attention (SWA)
& Shifts the window partition by half a window to exchange information across neighboring regions. \\

\rowcolor{gray!10}
Cached linear attention ($L$)
& Compresses the observation history into a reusable key--value summary for repeated queries. \\

Gated Delta Network (GDN)
& Maintains a recurrent state across frames for temporal history modeling. \\

\rowcolor{gray!10}
Full attention
& Provides global mixing in the final anchor block to preserve a direct long-range communication path. \\

\bottomrule
\end{tabularx}
\end{table}

\paragraph{Local target mixing}
Cached-Linear replaces global target self-attention with local window and
shifted-window attention. Window attention performs standard Softmax
self-attention only among the 64 tokens within each local window. For window
partition $P_r$,
\begin{equation}
\operatorname{WAttn}(X)_r
=
\operatorname{softmax}(Q_rK_r^\top/\sqrt d)V_r,
\qquad
X_r=P_r(X).
\label{eq:window-attention}
\end{equation}
For $N$ target tokens and window width $w$, this reduces the attention
computation from $O(N^2d)$ to $O(Nw^2d)$. Fixed local windows, however, limit
communication across window boundaries. We therefore use shifted-window
attention (SWA), which cyclically shifts the partition by half a window and
applies a boundary mask $\mathcal M$:
\begin{equation}
\operatorname{SWA}(X)
=
P_{w/2}^{-1}
\operatorname{WAttn}_{\mathcal M}(P_{w/2}X).
\label{eq:shifted-window}
\end{equation}
Alternating local and shifted-window mixing preserves exact local attention
while allowing information to propagate across neighboring regions.

\paragraph{Cached linear history access}
The observation history is unchanged across flow-integration steps and across
candidate actions evaluated from the same context. Cached-Linear exploits this
reuse by summarizing the history once and sharing the resulting cache across
target queries. Using the positive feature map
$\phi(x)=\operatorname{ELU}(x)+1$, history keys $K$ and values $V$ are
compressed into
\begin{equation}
\resizebox{0.98\columnwidth}{!}{$\displaystyle
M=\phi(K)^\top V,\qquad
u=\sum_j\phi(K_j),\qquad
\phi(x)=\operatorname{ELU}(x)+1.$}
\label{eq:linear-cache}
\end{equation}
A new target query $q$ reads only the cached pair $(M,u)$:
\begin{equation}
\operatorname{LinAttn}(q)
=
\frac{\phi(q)M}{\phi(q)u+\varepsilon}.
\label{eq:linear}
\end{equation}
The cache is constructed once with cost linear in history length, and each
subsequent query avoids materializing the full query--history score matrix.
This directly amortizes history processing across repeated queries. Because the
kernelized summary does not preserve every pairwise interaction represented by
full Softmax attention, Section~\ref{sec:results} evaluates the resulting
prediction-quality and efficiency trade-off.

\paragraph{Balanced GDN recurrent history model}
Cached-Linear represents the observation history through a reusable additive
summary. Balanced GDN augments this design with frame-wise recurrent memory,
allowing temporal information to be updated as the history is processed.
Specifically, one self-mixing block in each four-block group is replaced with
a Gated Delta Network. The remaining blocks combine local attention,
shifted-window attention, cached linear cross-attention, and a final global
attention anchor, as illustrated in
Figure~\ref{fig:gdn-architecture}.

Let $P=256$ denote the number of spatial tokens per frame,
$H_a=12$ the number of heads, and $d_h=64$ the head width. For each batch
element and attention head, the projected frame matrices
$Q_t,K_t,V_t\in\mathbb R^{d_h\times P}$ store tokens as columns. Spatial RoPE
produces $\bar Q_t$ and $\bar K_t$ with the same shape. The write gate
$\beta_t\in(0,1)^{1\times P}$ operates token-wise and per head, while the
retention gate $\alpha_t\in(0,1]$ is a scalar for each frame and head whose
input is the mean of the $P$ frame tokens. Define
$B_t=K_t\odot\beta_t$, $\bar B_t=\bar K_t\odot\beta_t$, and
$U_t=V_t\odot\beta_t$, all in $\mathbb R^{d_h\times P}$.

The recurrent matrix $S_t\in\mathbb R^{d_h\times d_h}$ is updated as
\begin{equation}
S_t
=
\alpha_t
\left[
S_{t-1}-(S_{t-1}\bar B_t)\bar K_t^\top
\right]
+
U_t\bar K_t^\top.
\label{eq:gdn-state}
\end{equation}
The normalization state $r_t\in\mathbb R^{d_h}$ and output $Y_t$ are
\begin{equation}
r_t
=
\alpha_t
\left[
r_{t-1}-B_t(K_t^\top r_{t-1})
\right]
+
B_t\mathbf 1_P,
\qquad
Y_t
=
\frac{S_t\bar Q_t}{r_t^\top Q_t+\varepsilon}.
\label{eq:gdn-normalizer}
\end{equation}
All $P$ token contributions within a frame are evaluated in parallel, while
frames are scanned in temporal order. An output gate projects
$Y_t\in\mathbb R^{d_h\times P}$ back to the model width. The recurrent states
require $O(d_h^2+d_h)$ memory per head and therefore do not grow with context
length. Unlike the additive cache used by Cached-Linear, the learned GDN update
can replace stale information as new frames are processed.

The 12 mixer blocks are organized into three groups of four. Within group $j$,
the self-mixers and cross-mixers are
\begin{equation}
\resizebox{0.98\columnwidth}{!}{$\displaystyle
\mathcal S_j=(\mathrm{GDN},W,W,A_j),\quad
\mathcal C_j=(L,L,L,\mathrm{SWA}),\quad
A_{1,2}=\mathrm{SWA},\quad A_3=\mathrm{Full}.$}
\label{eq:gdn-schedule}
\end{equation}
In this schedule, GDN provides frame-wise temporal memory, window attention
preserves local spatial interactions, shifted-window attention exchanges
information across neighboring regions, and cached linear attention enables
repeated history reuse. The final full-attention block retains a direct global
communication path.

\paragraph{Auxiliary iREPA spatial alignment}
We additionally evaluate an auxiliary Balanced GDN + iREPA variant. This
variant keeps the tokenizer, decoder, and history-mixer schedule unchanged.
During training, a $3\times3$ convolution projects one intermediate dynamics
feature to the normalized clean visual-feature target, and we add the iREPA
alignment loss~\cite{singh2025irepa}. Only one feature level is used, without
fusion across multiple backbone blocks. The alignment branch is removed at
inference.

Global-Softmax, Cached-Linear, Balanced GDN, and Balanced GDN + iREPA are
treated as complete model configurations in the component study. The iREPA
variant modifies the training objective but does not change the history-mixer
architecture.

\section{Experimental Design}
\label{sec:protocol}

\subsection{Evaluation Overview}
Table~\ref{tab:xuechu-study-map} lists the training budgets, datasets, sample
counts, and evaluated horizons. Each quality comparison uses the same
decoded-image protocol and keeps systems timing separate from prediction
quality.

\begin{table*}[!t]
\centering
\caption{Experimental design and evaluation map.}
\label{tab:xuechu-study-map}
\footnotesize
\renewcommand{\arraystretch}{1.15}
\setlength{\tabcolsep}{5pt}

\begin{tabularx}{\textwidth}{
    @{}
    >{\raggedright\arraybackslash\bfseries}p{0.19\textwidth}
    >{\raggedright\arraybackslash}p{0.15\textwidth}
    >{\raggedright\arraybackslash}p{0.20\textwidth}
    >{\raggedright\arraybackslash}p{0.10\textwidth}
    >{\raggedright\arraybackslash}X
    @{}
}
\toprule
Study & Representation & History mixer & Train & Evaluation \\
\midrule

Study I: Representation
& 5 states
& Global-Softmax
& 10 epochs
& 16\,s: 598/199/42;\quad 32\,s: 150/112/37 \\

Main Experiments
& RAE, PAE, V-JEPA
& 7 models
& 50 epochs
& Direct and rollout @ 16/32 \\

Ablation I: Components
& RAE, PAE, V-JEPA
& $3\times4$ configurations
& 10 epochs
& Direct and rollout @ 16/32 \\

Ablation II: Systems
& RAE-B
& Softmax / Linear / GDN
& Systems
& 8--128 frames; 1 or 120 queries \\

\bottomrule
\end{tabularx}

\vspace{0pt}

\end{table*}

\subsection{Data and Evaluation Sets}
Training and evaluation use RECON~\cite{shah2021rapid}, SACSoN
~\cite{hirose2023sacson}, and SCAND~\cite{karnan2022scand}. For each epoch, we
draw 62,813 observations per dataset to match the smallest source and
interleave the balanced streams. Each observation supplies four
future targets. With 48 observations per rank on two ranks, one optimizer step
contains 96 observations and 384 conditional transitions. Direct prediction samples each requested endpoint from real
history. Rollout advances at 4 frames per second and returns predicted latents
to the context without intermediate decode/re-encode. Paired comparisons reuse
sample identifiers and initial diffusion noise; all models use a 50-step Euler
solver.

The reference tasks follow RAE-NWM~\cite{zhang2026raenwm}. Direct prediction
generates the SACSoN Direct@4 and Direct@16 endpoints independently from real
history. Trajectory prediction covers two seconds in eight steps and searches
120 candidates with top-3 selection and one CEM update. The nominal complete
held-out sets contain 199 SACSoN, 598 RECON, and 42 SCAND trajectories. This is an open-loop
prediction test, not a closed-loop navigation evaluation.

\subsection{Training Configuration}
The shared backbone is a 12-block CDiT-B/2 of width 768 with 12 heads and a
$16\times16$ target grid, followed by a two-block, width-2,048, 16-head
prediction head. The optimizer is AdamW with
$\beta=(0.9,0.999)$, gradient-norm clipping at 1.0, and a linear learning-rate
schedule from $2\times10^{-4}$ to $2\times10^{-6}$. Training uses BF16, seed
42, deterministic sampling, and EMA checkpoints. Table
~\ref{tab:xuechu-study-map} lists the budgets used by each experiment.

The Global-Softmax and Cached-Linear RAE pair contains 364.24M trainable
parameters. For the cross-representation Balanced GDN study, we adjust the MLP ratio to
keep each bundle near 350M parameters despite different native channel widths
and the iREPA projector. 

\subsection{Metrics, Aggregation, and Uncertainty}
DINO distance is the primary decoded endpoint metric. LPIPS measures deep
perceptual similarity~\cite{zhang2018unreasonable}; DreamSim uses an ensemble
trained for human similarity judgments~\cite{fu2023dreamsim}; PSNR and SSIM
retain pixel-level information. Lower is better for DINO, LPIPS, and DreamSim;
higher is better for PSNR and SSIM.

For metric $m$, dataset $s$, and its $n_s$ held-out trajectories, we average
within each dataset and give the three datasets equal weight:
\begin{equation}
\bar m=\frac{1}{3}\sum_{s=1}^{3}\frac{1}{n_s}\sum_{i=1}^{n_s}m_{s,i}.
\label{eq:dataset-macro}
\end{equation}
We summarize the four endpoints as
\begin{equation}
\operatorname{Avg4}(m)=\frac{1}{4}
\sum_{q\in\{D,R\}}\sum_{h\in\{16,32\}}\bar m_{q,h},
\label{eq:avg4}
\end{equation}
where $D$ and $R$ denote direct prediction and rollout. Cross-model values use
one training seed. Episode-paired bootstrap intervals resample trajectories
within each dataset for the Global-Softmax/Cached-Linear RAE pair. They measure
evaluation variation, not variation from retraining.

\subsection{Conditioning and Control Diagnostics}
We corrupt the conditioning input by zeroing, shuffling, or sign-flipping the
action, and by replacing the requested horizon with a mismatched one. We
average the resulting increase in DINO distance as a measure of conditioning
sensitivity. An episode-disjoint linear probe from frozen states to relative
planar motion measures accessible motion information. These diagnostics test
whether a model reacts to its conditions.

\subsection{Joint Selection Protocol}
Table~\ref{tab:xuechu-study-map} is the paper's reference for training
budgets and evidence roles. The visual representation research compares five frozen states
with Global-Softmax. We then retain the three representation that lead one control
objective: PAE-L for reconstruction, RAE-B for direct prediction, and V-JEPA
for rollout. The main subset contains the RAE-B Global-Softmax and
Cached-Linear anchors, Balanced GDN for all three retained representation,
and the PAE-L/V-JEPA iREPA auxiliaries.  The $3\times4$ matrix is an exploratory
component study.

The primary systems tables compare three history-processing architectures—\emph{Global-Softmax}, \emph{Cached-Linear}, and \emph{Balanced GDN}—abbreviated as \emph{Softmax}, \emph{Linear}, and \emph{GDN}, respectively. Checkpoint-matched cost claims use only
benchmarks with the evaluated checkpoint's operator schedule.  Other fixed-shape schedules remain in a separate
exploratory screen. Results follow this selection order: visual-representation
control, main quality comparison, component ablation, and systems efficiency.
The first three groups each report their own direct-prediction and open-loop
trajectory results.

\begin{table}[!t]
\centering

\caption{\textbf{Study I: visual representation and reconstruction DINO.}}
\label{tab:representation-interfaces}
\footnotesize
\setlength{\tabcolsep}{3.2pt}
\renewcommand{\arraystretch}{1.08}

\begin{tabular}{@{}l l r r r@{}}
\toprule
Interface
& Frozen source
& Channels
& Values/frame
& Re. DINO $\downarrow$ \\
\midrule

RAE-B
& DINOv2-B/C768
& 768
& 196,608
& 0.147 \\

RAE-L
& DINOv2-L/C1024
& 1,024
& 262,144
& 0.159 \\

PAE-L
& DINOv2-L/C32
& 32
& 8,192
& \best{0.069} \\

Cosmos
& Cosmos-CI16
& 16
& 4,096
& 0.125 \\

V-JEPA
& V-JEPA 2.1 ViT-L
& 96
& 24,576
& 0.335 \\

\bottomrule
\end{tabular}

\end{table}

\vspace{0pt}

\section{Results}
\label{sec:results}

We organize the results around four questions: which visual representation works best for different prediction objectives, how history mixers affect prediction quality, how representation and mixer choices interact, and how efficiently each method reuses a shared observation history. We also report open-loop CEM trajectory results alongside the corresponding prediction results.

\subsection{Study I: Visual Representation Research}
\label{sec:study1-results}

\begin{figure}[t]
    \centering
    \includegraphics[width=\columnwidth]{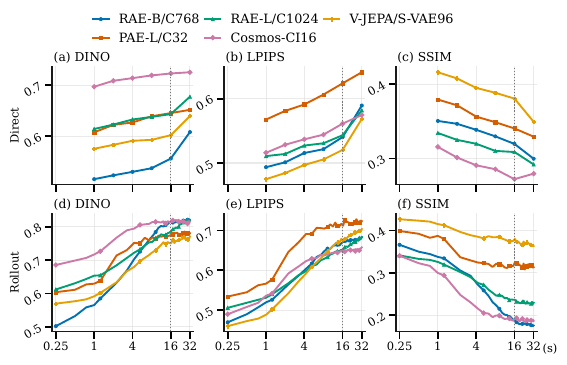}
    \vspace{-4pt}
    \caption{\textbf{Study I: Visual-Representation Horizons.} Horizon-wise decoded quality for all five Representation under Global-Softmax. Rows show direct prediction and 4-fps autoregressive rollout; columns show DINO, LPIPS, and SSIM. Lower is better for DINO/LPIPS and higher is better for SSIM. }
    \label{fig:representation-horizons}
    \vspace{0pt}
\end{figure}

\subsubsection{Different Prediction Objectives Favor Different Visual Representations}


We first compare the five visual representation under the same Global-Softmax
setting. The results show that no single representation performs best across
all prediction objectives.

For reconstruction, PAE-L achieves the lowest DINO score of 0.069. For direct
prediction, RAE-B performs best, with DINO scores of 0.574 at Direct@16 and
0.608 at Direct@32. For long-horizon rollout, V-JEPA performs best, reaching
0.761 at Rollout@16 and 0.771 at Rollout@32. These results show that good
reconstruction does not necessarily lead to good future prediction. Figure~\ref{fig:representation-horizons} shows how this difference develops
over time. RAE-B performs better at short rollout horizons, but its error grows
more as the prediction horizon increases. V-JEPA starts behind RAE-B but
overtakes it after 3.25~s. This suggests that a representation that works well
when predicting from real observations may not remain the best choice when the
model repeatedly uses its own predictions. Overall, the preferred representation changes with the prediction objective:
PAE-L is strongest for reconstruction, RAE-B for direct prediction, and
V-JEPA for long-horizon rollout under this matched control.

\subsubsection{Representation Rankings Vary Across Metrics and Trajectory Prediction}


The additional image metrics give a more detailed view of the same comparison.
Table~\ref{tab:reference-study1-direct} shows that RAE-B achieves the lowest
DreamSim and DINO, while V-JEPA achieves the lowest LPIPS and FID-199.
Different metrics therefore favor different aspects of the predicted images.
We use DINO as the primary metric and report the others as complementary
measures. The open-loop trajectory results also vary across datasets.
Table~\ref{tab:reference-study1-planning} shows that RAE-B performs best on
SACSoN and gives the lowest ATE on RECON. RAE-L gives the lowest RPE on
RECON, while V-JEPA performs best on SCAND. Therefore, the representation that
performs best on image prediction is not always the best one for trajectory
prediction across different datasets. Together, these results show that visual-representation rankings depend not only on
the prediction objective, but also on the evaluation metric and dataset.

\subsubsection{Visual Representations Respond to Motion Conditions}


Finally, we test whether the models use the action and prediction-horizon
conditions. When these inputs are removed, shuffled, or changed, prediction
quality becomes worse on average for all five visual representation. RAE-B is the
most sensitive to these changes, while Cosmos is the least sensitive. This
shows that the models use the conditioning information during prediction.

We also test whether relative robot motion can be recovered directly from the
frozen visual features using a linear probe. All five representation obtain
negative macro $R^2$, indicating that none of them contains a clear linear
representation of relative SE(2) motion. V-JEPA has the highest Pearson
correlation, but the value remains small. These results show that the models respond to action and horizon information,
but the frozen visual features do not provide a simple linear encoding of
robot motion.

\vspace{-5pt}
\begin{table}[!t]
\centering
\caption{\textbf{Study I: direct prediction.} One-step SACSoN Direct@4 and Direct@16 on 199 held-out trajectories.}
\label{tab:reference-study1-direct}
\footnotesize
\renewcommand{\arraystretch}{1.05}
\setlength{\tabcolsep}{3pt}

\begin{tabular}{@{}l cc cc cc cc@{}}
\toprule
& \multicolumn{2}{c}{LPIPS $\downarrow$}
& \multicolumn{2}{c}{DreamSim $\downarrow$}
& \multicolumn{2}{c}{DINO $\downarrow$}
& \multicolumn{2}{c}{FID-199 $\downarrow$} \\
\cmidrule(lr){2-3}
\cmidrule(lr){4-5}
\cmidrule(lr){6-7}
\cmidrule(lr){8-9}

Model
& 4s & 16s
& 4s & 16s
& 4s & 16s
& 4s & 16s \\
\midrule

RAE-B
& \second{0.492} & \second{0.529}
& \best{0.335} & \best{0.377}
& \best{0.497} & \best{0.529}
& \second{156.00} & \second{170.36} \\

PAE-L
& 0.583 & 0.605
& 0.450 & 0.472
& 0.605 & 0.625
& 171.59 & 181.79 \\

RAE-L
& 0.505 & 0.529
& 0.420 & 0.456
& 0.626 & 0.655
& 209.55 & 223.86 \\

Cosmos
& 0.550 & 0.570
& 0.528 & 0.546
& 0.722 & 0.738
& 247.27 & 250.87 \\

V-JEPA
& \best{0.466} & \best{0.499}
& \second{0.353} & \second{0.386}
& \second{0.571} & \second{0.593}
& \best{133.82} & \best{145.96} \\

\bottomrule
\end{tabular}
\end{table}

\begin{table}[!t]
\centering
\caption{\textbf{Study I: open-loop trajectory prediction.} Two-second, eight-step trajectory prediction on held-out indices, using CEM-120, DINO-token scoring, and Euler-50 generation.}
\label{tab:reference-study1-planning}
\footnotesize
\renewcommand{\arraystretch}{1.08}
\setlength{\tabcolsep}{3.5pt}

\begin{tabular}{@{}l cc cc cc@{}}
\toprule
& \multicolumn{2}{c}{\textbf{SACSoN}}
& \multicolumn{2}{c}{\textbf{RECON}}
& \multicolumn{2}{c}{\textbf{SCAND}} \\
\cmidrule(lr){2-3}
\cmidrule(lr){4-5}
\cmidrule(l){6-7}

Model
& ATE $\downarrow$ & RPE $\downarrow$
& ATE $\downarrow$ & RPE $\downarrow$
& ATE $\downarrow$ & RPE $\downarrow$ \\
\midrule

RAE-B/C768
& \textbf{3.668} & \textbf{0.858}
& \textbf{1.644} & 0.448
& 1.210 & 0.326 \\

PAE-L/C32
& 4.016 & 0.907
& 1.765 & 0.476
& 1.170 & 0.316 \\

RAE-L/C1024
& 3.916 & 0.914
& 1.689 & \textbf{0.443}
& 1.214 & 0.330 \\

Cosmos-CI16
& 4.369 & 0.993
& 1.803 & 0.468
& 1.154 & 0.316 \\

V-JEPA/S-VAE96
& 3.885 & 0.909
& 1.754 & 0.448
& \textbf{0.968} & \textbf{0.300} \\

\bottomrule
\end{tabular}
\end{table}

\begin{table}[!t]
\centering
\caption{\textbf{Study I: conditioning sensitivity and frozen motion probes.}}
\label{tab:yier-action-probe-atlas}
\footnotesize
\renewcommand{\arraystretch}{1.05}
\setlength{\tabcolsep}{3pt}

\begin{tabular}{@{}lccccc@{}}
\toprule
\multicolumn{6}{l}{\textit{(a) Conditioning corruption: DINO degradation}} \\
\midrule
Model & Macro & RECON & SACSoN & SCAND & Clean DINO \\
\midrule

RAE-B
& \textbf{+0.0369}
& \textbf{+0.0255}
& \textbf{+0.0442}
& \textbf{+0.0411}
& 0.5742 \\

PAE-L
& +0.0075
& +0.0003
& +0.0068
& +0.0153
& 0.6520 \\

RAE-L
& +0.0222
& +0.0186
& +0.0231
& +0.0250
& 0.6607 \\

Cosmos
& +0.0071
& -0.0007
& +0.0064
& +0.0157
& 0.7355 \\

V-JEPA
& +0.0227
& +0.0157
& +0.0301
& +0.0223
& 0.6181 \\

\midrule
\multicolumn{5}{l}{\textit{(b) Frozen relative-SE(2) probe}}\\
\midrule
Model & Pearson $r$ & Macro $R^2$ & RECON $R^2$ & SACSoN $R^2$ & SCAND $R^2$\\
\midrule

RAE-B
& +0.013 & -3.931 & -2.369 & -3.178 & -6.246\\

PAE-L
& +0.073 & -0.164 & -0.055 & -0.135 & -0.303\\

RAE-L
& +0.052 & -2.863 & -1.053 & -2.693 & -4.842\\

Cosmos
& +0.021 & \textbf{-0.123} & \textbf{-0.035} & \textbf{-0.121} & \textbf{-0.212}\\

V-JEPA
& \textbf{+0.103} & -0.327 & -0.329 & -0.172 & -0.479\\

\bottomrule
\end{tabular}
\end{table}

\subsection{Main Experiment: Representation--Mixer Trade-offs}
\label{sec:main50-results}

We next compare seven selected configurations built from the retained visual
representation and history-processing designs. This comparison focuses on three
questions: how the history mixer changes direct prediction and rollout,
whether the preferred configuration changes across metrics and datasets, and
whether better image prediction also leads to better open-loop trajectory
prediction.

\begin{figure}[t]
    \centering
    \includegraphics[width=0.90\columnwidth]{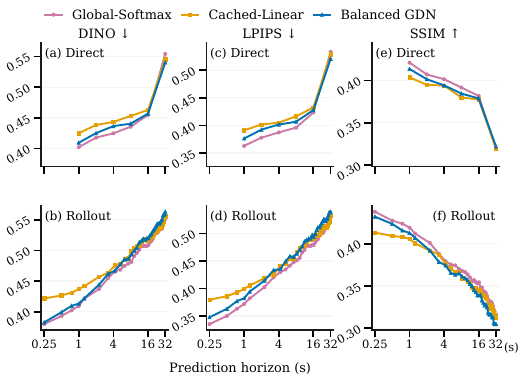}
    \vspace{-10pt}
    \caption{\textbf{Main Experiment: RAE Horizon Quality.} Macro-averaged
    Direct (top) and autoregressive Rollout (bottom) curves across RECON,
    SACSoN, and SCAND. Columns report DINO, LPIPS, and SSIM.}
    \label{fig:rae-global-cached-horizons}
    \vspace{-15pt}
\end{figure}

\subsubsection{History Mixers Affect Direct Prediction and Rollout Differently}

We first compare Global-Softmax, Cached-Linear, and Balanced GDN using RAE-B.
Their overall DINO results are close, but they behave differently across
prediction modes. Balanced GDN gives the lowest Direct@32 DINO, improving from
0.554 with Global-Softmax to 0.540. However, this improvement does not carry
over to rollout: Rollout@16 increases from 0.530 to 0.542 and Rollout@32 from
0.554 to 0.563. Figure~\ref{fig:rae-global-cached-horizons} shows the same pattern across
prediction horizons. The three methods remain close for direct prediction,
while their differences become more visible during rollout. This suggests
that changing the history mixer can improve some direct-prediction endpoints
without necessarily improving stability when predicted states are fed back
into the model. The overall Avg-4 DINO scores are also very similar: 0.5300 for
Global-Softmax, 0.5310 for Cached-Linear, and 0.5314 for Balanced GDN.
Therefore, none of the three history mixers gives a clear quality advantage
across all four endpoints. Their main differences appear in where the
prediction error occurs rather than in the overall average.


\begin{table*}[!t]
\centering
\caption{\textbf{Main experiment: endpoint results.}
Macro-averages across RECON, SACSoN, and SCAND.}
\label{tab:xuechu-e50-endpoints}

\scriptsize
\renewcommand{\arraystretch}{1.05}
\setlength{\tabcolsep}{1.8pt}

\resizebox{\textwidth}{!}{%
\begin{tabular}{@{}ll*{20}{c}@{}}
\toprule
&
&
\multicolumn{5}{c}{Direct@16}
&
\multicolumn{5}{c}{Rollout@16}
&
\multicolumn{5}{c}{Direct@32}
&
\multicolumn{5}{c}{Rollout@32}
\\

\cmidrule(lr){3-7}
\cmidrule(lr){8-12}
\cmidrule(lr){13-17}
\cmidrule(l){18-22}

State & Mixer
& DINO & LPIPS & DreamSim & PSNR & SSIM
& DINO & LPIPS & DreamSim & PSNR & SSIM
& DINO & LPIPS & DreamSim & PSNR & SSIM
& DINO & LPIPS & DreamSim & PSNR & SSIM
\\
\midrule

RAE & Softmax
& 0.482 & 0.451 & 0.235 & \second{12.98} & 0.359
& \best{0.530} & \best{0.505} & \best{0.314} & \best{11.77} & 0.330
& 0.554 & 0.534 & 0.311 & 11.47 & 0.319
& \second{0.554} & \second{0.531} & \best{0.337} & \best{11.31} & 0.316
\\

RAE & Linear
& 0.483 & 0.456 & 0.241 & 12.91 & 0.356
& 0.539 & 0.515 & 0.321 & \second{11.72} & 0.323
& \second{0.546} & \second{0.529} & \second{0.306} & \second{11.49} & 0.319
& 0.556 & 0.532 & 0.347 & 11.11 & 0.313
\\

RAE & GDN
& 0.480 & 0.455 & 0.235 & 12.88 & 0.355
& 0.542 & 0.522 & 0.327 & 11.53 & 0.319
& \best{0.540} & \best{0.521} & \best{0.303} & \best{11.71} & 0.322
& 0.563 & 0.540 & 0.353 & 11.05 & 0.305
\\

PAE & GDN
& \best{0.478} & \second{0.436} & \second{0.231} & 12.86 & 0.375
& 0.535 & 0.508 & 0.319 & 11.50 & 0.332
& 0.567 & 0.538 & 0.334 & 11.20 & \second{0.333}
& 0.565 & 0.540 & 0.350 & 10.94 & 0.319
\\

PAE & GDN+iREPA
& \second{0.479} & \best{0.434} & \best{0.230} & \best{13.00} & 0.377
& \second{0.532} & \second{0.506} & \second{0.316} & 11.60 & 0.340
& 0.568 & 0.537 & 0.337 & 11.19 & 0.322
& \best{0.553} & \best{0.527} & \second{0.341} & \second{11.13} & \second{0.328}
\\

V-JEPA & GDN
& 0.539 & 0.482 & 0.314 & 12.86 & \best{0.387}
& 0.585 & 0.542 & 0.367 & 11.41 & \best{0.347}
& 0.604 & 0.557 & 0.382 & 11.10 & \best{0.341}
& 0.604 & 0.565 & 0.397 & 10.96 & \best{0.334}
\\

V-JEPA & GDN+iREPA
& 0.545 & 0.485 & 0.311 & 12.72 & \second{0.384}
& 0.588 & 0.548 & 0.374 & 11.38 & \second{0.343}
& 0.614 & 0.569 & 0.395 & 10.73 & 0.331
& 0.602 & 0.569 & 0.392 & 10.77 & 0.324
\\

\bottomrule
\end{tabular}%
}
\par \vspace{3pt}

\footnotesize
Softmax: Global-Softmax, Linear: Cached-Linear, GDN: Balanced GDN.

\end{table*}

\subsubsection{Preferred Configurations Vary Across Metrics and Datasets}

\begin{table}[!t]
\centering
\caption{\textbf{Main experiment: direct prediction.} One-step SACSoN Direct@4 and Direct@16 on 199 held-out trajectories.}
\label{tab:reference-main-direct}
\footnotesize
\renewcommand{\arraystretch}{1.05}
\setlength{\tabcolsep}{3pt}
\resizebox{\columnwidth}{!}{%

\begin{tabular}{@{}lcccccccc@{}}
\toprule
Model
& \multicolumn{2}{c}{LPIPS $\downarrow$}
& \multicolumn{2}{c}{DreamSim $\downarrow$}
& \multicolumn{2}{c}{DINO $\downarrow$}
& \multicolumn{2}{c}{FID $\downarrow$} \\
\cmidrule(lr){2-3}
\cmidrule(lr){4-5}
\cmidrule(lr){6-7}
\cmidrule(l){8-9}

& 4s & 16s
& 4s & 16s
& 4s & 16s
& 4s & 16s \\
\midrule

RAE-Softmax
& 0.339 & 0.382
& 0.165 & 0.198
& 0.359 & 0.401
& 63.43 & 69.61 \\

RAE-Linear
& 0.344 & 0.381
& 0.167 & 0.194
& 0.368 & 0.396
& 62.98 & 68.55 \\

RAE-GDN
& 0.351 & 0.384
& 0.170 & 0.193
& 0.365 & 0.396
& 62.70 & 68.56 \\

PAE-GDN
& \second{0.304} & \second{0.354}
& \second{0.139} & \best{0.172}
& \second{0.343} & \second{0.382}
& \best{58.14} & \second{64.55} \\

PAE-GDN+iREPA
& \best{0.301} & \best{0.352}
& \best{0.137} & \second{0.173}
& \best{0.341} & \best{0.381}
& \second{58.17} & \best{64.37} \\

VJEPA-GDN
& 0.383 & 0.421
& 0.240 & 0.263
& 0.458 & 0.479
& 83.74 & 92.74 \\

VJEPA-GDN+iREPA
& 0.384 & 0.420
& 0.239 & 0.265
& 0.460 & 0.485
& 84.63 & 93.39 \\

\bottomrule
\end{tabular}

}

\end{table}

\begin{figure}[t]
    \centering
    \vspace{-5pt}
    \includegraphics[width=\columnwidth]{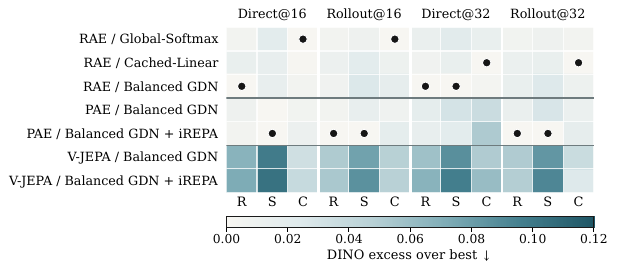}
    \vspace{-20pt}
       \caption{\textbf{Main Experiment: Seven-Model Subset by Dataset.} Each cell
    shows DINO excess over the best row for one dataset and
    endpoint; black dots mark the column winners.  R/S/C denote
    RECON/SACSoN/SCAND.}
    \label{fig:e50-dataset-robustness}
    \vspace{-10pt}
\end{figure}

The seven configurations also behave differently across visual representation and
image metrics. Table~\ref{tab:xuechu-e50-endpoints} shows that RAE /
Global-Softmax gives the best Avg-4 DINO, DreamSim, and PSNR, while PAE /
Balanced GDN + iREPA gives the best LPIPS and V-JEPA / Balanced GDN gives the
best SSIM. Thus, no configuration is best under every image-quality metric. The SACSoN direct-prediction results show another change in ranking.
Table~\ref{tab:reference-main-direct} shows that the two PAE-L configurations
occupy the best two positions across the direct-prediction columns. This
differs from the macro DINO results across RECON, SACSoN, and SCAND, where
RAE-based configurations are generally stronger. Figure~\ref{fig:e50-dataset-robustness} shows that the preferred configuration
also changes across datasets. SCAND selects an RAE-B configuration in all four
DINO endpoints, while PAE / Balanced GDN + iREPA leads several RECON and
SACSoN endpoints. The macro average gives equal weight to the three datasets,
so these dataset-level differences are not hidden by the larger RECON set. Together, these results show that the preferred configuration depends on both
the evaluation metric and the dataset. A single model ranking therefore does
not describe all prediction settings well.

\subsubsection{Image-Prediction Gains Do Not Necessarily Translate to Better Trajectory Prediction}

Finally, we compare the seven configurations on open-loop CEM trajectory
prediction. Table~\ref{tab:main50-planning} shows a much clearer result than
the image metrics: RAE / Global-Softmax achieves the lowest ATE and RPE on all
three datasets. It reaches 3.259/0.777 on SACSoN, 1.468/0.401 on RECON, and
1.021/0.285 on SCAND. This result is important because the configurations that perform best on
individual image metrics do not necessarily perform best on trajectory
prediction. For example, PAE-based configurations are strong on several
direct-prediction metrics, while V-JEPA-based configurations can perform well
on SSIM, but neither gives the best trajectory results in this comparison.
No Balanced GDN configuration wins a trajectory column. The trajectory results therefore provide a different view from the decoded
image metrics. Improvements on a particular image-prediction metric do not
automatically translate into better open-loop trajectory prediction. In this
comparison, RAE / Global-Softmax provides the most consistent trajectory
performance across all three datasets.

\begin{table}[!t]
\centering
\caption{\textbf{Main experiment: open-loop trajectory prediction.} Two-second, eight-step planning. CEM uses 120 candidates, DINO-token scoring, and Euler-50 generation.}
\label{tab:main50-planning}
\footnotesize
\renewcommand{\arraystretch}{1.05}
\setlength{\tabcolsep}{3pt}

\begin{tabular}{@{}lcccccc@{}}
\toprule
& \multicolumn{2}{c}{SACSoN}
& \multicolumn{2}{c}{RECON}
& \multicolumn{2}{c}{SCAND} \\
\cmidrule(lr){2-3}
\cmidrule(lr){4-5}
\cmidrule(lr){6-7}

Model
& ATE $\downarrow$ & RPE $\downarrow$
& ATE $\downarrow$ & RPE $\downarrow$
& ATE $\downarrow$ & RPE $\downarrow$ \\
\midrule

RAE-Softmax
& \best{3.259} & \best{0.777}
& \best{1.468} & \best{0.401}
& \best{1.021} & \best{0.285} \\

RAE-Linear
& 3.458 & 0.807
& 1.664 & 0.426
& \second{1.083} & 0.302 \\

RAE-GDN
& \second{3.325} & 0.786
& \second{1.553} & \second{0.414}
& 1.276 & 0.336 \\

PAE-GDN
& 3.614 & 0.836
& 1.668 & 0.432
& 1.218 & 0.336 \\

PAE-GDN+iREPA
& 3.578 & 0.818
& 1.633 & 0.424
& 1.220 & 0.325 \\

VJEPA-GDN
& 3.453 & 0.796
& 1.610 & 0.417
& 1.121 & \second{0.298} \\

VJEPA-GDN+iREPA
& 3.330 & \second{0.786}
& 1.621 & 0.420
& 1.245 & 0.327 \\

\bottomrule
\end{tabular}


\end{table}

\subsection{Ablation I: Component Bundles}
\label{sec:ablation-results}

Table~\ref{tab:xuechu-component-endpoints} shows that the preferred history
mixer depends on the visual representation. Balanced GDN gives the lowest
Avg-4 DINO for RAE-B and PAE-L, while Cached-Linear gives the best result for
V-JEPA. For PAE-L, Balanced GDN performs better on direct prediction, whereas
Cached-Linear performs better on rollout. These results indicate that no
single history mixer is consistently best across representations and
prediction modes.

Adding iREPA does not improve prediction quality in this ablation. Balanced
GDN + iREPA produces higher DINO than Balanced GDN in all 12 comparisons,
showing no clear benefit from the additional alignment objective under this
setting. The SACSoN direct-prediction results in
Table~\ref{tab:reference-ablation-direct} show a similar pattern. RAE-B with
Balanced GDN gives the lowest Direct@4 DINO, while RAE-B with Cached-Linear
gives the lowest Direct@16 DINO. V-JEPA with Cached-Linear gives the lowest
FID-199 at both horizons. The preferred configuration therefore also changes
with the prediction horizon and evaluation metric.

Finally, the conditioning results show that all evaluated models respond to
the action and prediction-horizon inputs on average. The mean DINO increase
ranges from 0.0034 to 0.0318 when these conditions are changed. Although one
PAE-L case does not follow this trend, the overall results indicate that the
models use the conditioning information during prediction. Table~\ref{tab:reference-ablation-planning} applies the same open-loop CEM-120
task to the component matrix. 



\begin{table}[!t]
\centering
\caption{\textbf{Ablation I: architecture matrix.} The $3\times4$
endpoint and Avg-4 DINO results compare complete bundles.}
\label{tab:xuechu-component-endpoints}
\footnotesize
\renewcommand{\arraystretch}{1.05}
\setlength{\tabcolsep}{3pt}
\resizebox{\columnwidth}{!}{%
\begin{tabular}{@{}lccccc@{}}
\toprule
Model & Direct@16 & Rollout@16 & Direct@32 & Rollout@32 & Avg-4 \\
\midrule

RAE-Softmax
& 0.574 & 0.813 & 0.608 & 0.820 & 0.704 \\

RAE-Linear
& \second{0.550} & \second{0.667}
& \second{0.571} & \second{0.692}
& \second{0.620} \\

RAE-GDN
& \best{0.543} & \best{0.656}
& \best{0.567} & \best{0.677}
& \best{0.611} \\

RAE-GDN+iREPA
& 0.564 & 0.681 & 0.581 & 0.702 & 0.632 \\

\midrule

PAE-Softmax
& 0.646 & 0.815 & 0.652 & 0.816 & 0.732 \\

PAE-Linear
& \second{0.628} & \best{0.736}
& \second{0.635} & \best{0.747}
& \second{0.687} \\

PAE-GDN
& \best{0.622} & \second{0.745}
& \best{0.618} & \second{0.758}
& \best{0.686} \\

PAE-GDN+iREPA
& 0.659 & 0.763 & 0.656 & 0.759 & 0.709 \\

\midrule

V-JEPA-Softmax
& 0.618 & 0.761 & 0.639 & 0.771 & 0.697 \\

V-JEPA-Linear
& \second{0.612} & \best{0.725}
& \second{0.638} & \best{0.729}
& \best{0.676} \\

V-JEPA-GDN
& \best{0.608} & \second{0.735}
& \best{0.633} & \second{0.760}
& \second{0.684} \\

V-JEPA-GDN+iREPA
& 0.621 & 0.752 & 0.645 & 0.767 & 0.696 \\

\bottomrule
\end{tabular}
}
\end{table}

\begin{table}[!t]
\centering
\caption{\textbf{Ablation I: direct prediction.} One-step SACSoN Direct@4 and Direct@16 on 199 held-out trajectories.}
\label{tab:reference-ablation-direct}
\footnotesize
\renewcommand{\arraystretch}{1.05}
\setlength{\tabcolsep}{2.2pt}

\resizebox{\columnwidth}{!}{%
\begin{tabular}{@{}lcccccccc@{}}
\toprule
Model
& \multicolumn{2}{c}{LPIPS $\downarrow$}
& \multicolumn{2}{c}{DreamSim $\downarrow$}
& \multicolumn{2}{c}{DINO $\downarrow$}
& \multicolumn{2}{c}{FID $\downarrow$} \\
\cmidrule(lr){2-3}
\cmidrule(lr){4-5}
\cmidrule(lr){6-7}
\cmidrule(l){8-9}
& 4s & 16s
& 4s & 16s
& 4s & 16s
& 4s & 16s \\
\midrule

RAE-Softmax
& 0.492 & 0.529
& 0.335 & 0.377
& 0.497 & 0.529
& 156.00 & 170.36 \\

RAE-Linear
& 0.483 & 0.508
& \second{0.307} & \best{0.340}
& \second{0.470} & \best{0.493}
& 135.11 & 147.78 \\

RAE-GDN
& 0.487 & 0.526
& \best{0.305} & \second{0.349}
& \best{0.460} & \second{0.497}
& 134.71 & 154.36 \\

RAE-GDN+iREPA
& 0.492 & 0.529
& 0.326 & 0.368
& 0.481 & 0.514
& 149.05 & 171.17 \\

\midrule

PAE-Softmax
& 0.570 & 0.590
& 0.437 & 0.449
& 0.609 & 0.615
& 161.57 & 159.14 \\

PAE-Linear
& 0.565 & 0.579
& 0.437 & 0.442
& 0.591 & 0.593
& 161.65 & 158.31 \\

PAE-GDN
& 0.562 & 0.587
& 0.425 & 0.447
& 0.583 & 0.600
& 165.84 & 168.66 \\

PAE-GDN+iREPA
& 0.577 & 0.599
& 0.449 & 0.463
& 0.611 & 0.617
& 171.32 & 175.88 \\

\midrule

V-JEPA-Softmax
& \best{0.466} & 0.499
& 0.353 & 0.386
& 0.571 & 0.593
& 133.82 & 145.96 \\

V-JEPA-Linear
& 0.471 & \best{0.494}
& 0.347 & 0.374
& 0.565 & 0.587
& \best{127.45} & \best{142.60} \\

V-JEPAGDN
& \second{0.469} & \second{0.498}
& 0.349 & 0.385
& 0.565 & 0.589
& \second{131.86} & \second{144.81} \\

V-JEPA-GDN+iREPA
& 0.482 & 0.519
& 0.367 & 0.405
& 0.571 & 0.598
& 132.11 & 149.51 \\

\bottomrule
\end{tabular}%
}

\end{table}

\begin{table}[!t]
\centering
\caption{\textbf{Ablation I: open-loop trajectory prediction.} Two-second, eight-step open-loop trajectory prediction on held-out indices, using CEM-120, top-3 selection, one update, DINO-token scoring, and Euler-50 generation.}
\label{tab:reference-ablation-planning}
\footnotesize
\renewcommand{\arraystretch}{1.05}
\setlength{\tabcolsep}{3pt}

\begin{tabular}{@{}lcccccc@{}}
\toprule
Model
& \multicolumn{2}{c}{SACSoN}
& \multicolumn{2}{c}{RECON}
& \multicolumn{2}{c}{SCAND} \\
\cmidrule(lr){2-3}
\cmidrule(lr){4-5}
\cmidrule(l){6-7}
& ATE $\downarrow$ & RPE $\downarrow$
& ATE $\downarrow$ & RPE $\downarrow$
& ATE $\downarrow$ & RPE $\downarrow$ \\
\midrule

RAE-Softmax
& \second{3.668} & \second{0.858}
& 1.644 & 0.448
& 1.210 & 0.326 \\

RAE-Linear
& 3.894 & 0.895
& 1.677 & 0.444
& 1.220 & 0.341 \\

RAE-GDN
& 3.816 & 0.896
& \second{1.591} & \second{0.436}
& 1.290 & 0.349 \\

RAE-GDN+iREPA
& 3.748 & 0.880
& \best{1.567} & \best{0.422}
& 1.227 & 0.343 \\

\midrule

PAE-Softmax
& 4.184 & 0.959
& 1.816 & 0.491
& 1.170 & 0.316 \\

PAE-Linear
& 3.802 & 0.882
& 1.764 & 0.483
& 1.200 & 0.324 \\

PAE-GDN
& 4.982 & 1.127
& 1.870 & 0.499
& 1.183 & 0.321 \\

PAE-GDN+iREPA
& 4.286 & 0.948
& 1.765 & 0.485
& 1.151 & 0.313 \\

\midrule

V-JEPA-Softmax
& 3.885 & 0.909
& 1.754 & 0.448
& 0.968 & 0.300 \\

V-JEPA-Linear
& \best{3.641} & \best{0.845}
& 1.763 & 0.453
& 0.962 & \best{0.295} \\

V-JEPA-GDN
& 3.769 & 0.868
& 1.767 & 0.455
& \second{0.957} & \second{0.297} \\

V-JEPA-GDN+iREPA
& 4.345 & 0.998
& 1.916 & 0.495
& \best{0.950} & 0.302 \\

\bottomrule
\end{tabular}

\end{table}

\begin{figure}[t]
    \centering
    \includegraphics[width=\columnwidth]{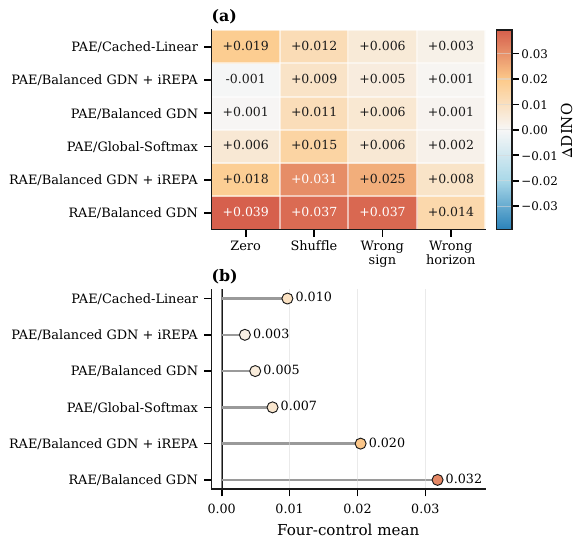}
    \vspace{-10pt}
    \caption{\textbf{Ablation I: Conditioning Controls.}
    Six-model subset. \textbf{(a)} DINO degradation under
    zero, shuffle, wrong-sign, and wrong-horizon inputs. \textbf{(b)}
    Four-control mean. Positive values indicate aggregate sensitivit.}
    \label{fig:conditioning-subset}
    \vspace{-10pt}
\end{figure}

\subsection{Ablation II: Systems Efficiency Test}
\subsubsection{Shared-History Efficiency and Quality Trade-off}

Global-Softmax is slightly faster for a single query with the standard
eight-frame history, but reusable history processing becomes substantially
more efficient when the same context is shared across many candidates.
With 64 history frames and 120 candidate queries, Cached-Linear reduces
latency from 51.67~s to 7.15~s and peak memory from 34.63~GiB to
5.35~GiB. At 128 frames, Global-Softmax runs out of memory, while
Cached-Linear remains nearly unchanged
(Table~\ref{tab:xuechu-systems-full}). Balanced GDN requires slightly more
computation than Cached-Linear but remains substantially more efficient
than Global-Softmax under shared-history reuse. It also improves selected
direct-prediction endpoints, including RAE-B Direct@32
(Table~\ref{tab:xuechu-e50-endpoints}). Overall, Cached-Linear provides
the strongest efficiency, while Balanced GDN offers a recurrent
quality--efficiency trade-off.

\subsubsection{Architecture Efficiency Comparison}

Table~\ref{tab:study2-architecture-screen} compares additional mixer
designs under a fixed workload on an RTX PRO 6000 Blackwell GPU.
Cached-Linear and Balanced GDN achieve similar query latency
(75.98~ms and 77.21~ms, respectively), both substantially lower than
Global-Softmax at 151.95~ms. Other GDN and MLA variants generally
introduce higher computational cost, supporting the balanced design as
an efficient recurrent alternative.

\begin{table}[!t]
\centering
\caption{\textbf{Ablation II: repeated-query scaling.} Cells report H100 BF16 latency (s) / peak memory
(GiB).}
\label{tab:xuechu-systems-full}
\footnotesize
\renewcommand{\arraystretch}{1.08}
\setlength{\tabcolsep}{4pt}
\begin{tabular}{@{}lccc@{}}
\toprule
Queries / frames & Global-Softmax & Cached-Linear & Balanced GDN \\
\midrule
Single / 8 & \best{0.58 / 2.10} & \second{0.60 / 2.13} & 0.78 / 2.19 \\
\midrule
120 / 8 & 11.84 / 7.50 & \best{7.15 / 5.29} & \second{7.41 / 5.40} \\
120 / 16 & 17.45 / 11.38 & \best{7.15 / 5.30} & \second{7.50 / 5.44} \\
120 / 32 & 28.81 / 19.13 & \best{7.15 / 5.31} & \second{7.68 / 5.51} \\
120 / 64 & 51.67 / 34.63 & \best{7.15 / 5.35} & \second{8.02 / 5.67} \\
120 / 128 & OOM / $>79.18$ & \best{7.19 / 5.55} & \second{8.75 / 5.97} \\
\bottomrule
\end{tabular}
\end{table}


\begin{table}[!t]
\centering
\caption{\textbf{Ablation II: architecture efficiency test.} BF16 model-core efficiency evaluation on one RTX PRO 6000 Blackwell GPU.}
\label{tab:study2-architecture-screen}
\footnotesize
\renewcommand{\arraystretch}{1.05}
\setlength{\tabcolsep}{3.5pt}
\resizebox{\columnwidth}{!}{%

\begin{tabular}{@{}lrrrr@{}}
\toprule
Architecture
& Train $\downarrow$ ms
& Memory $\downarrow$ GiB
& Prefill $\downarrow$ ms
& Query $\downarrow$ ms \\
\midrule

Global-Softmax
& 453.98 & 31.02 & -- & 151.95 \\

Cached-Linear
& \best{284.98} & \best{23.70}
& 12.73 & \best{75.98} \\

Linear + Cross-L3/MLA1
& 359.41 & 28.72
& \best{10.98} & 109.20 \\

Linear + Cross-L3/SWA1
& 293.94 & 24.44
& 15.07 & 77.71 \\

\midrule

Balanced GDN
& 292.66 & 24.76
& 13.52 & 77.21 \\

Dense Self-GDN
& 308.01 & 26.89
& 15.10 & 79.68 \\

Dual GDN/SWA
& 384.23 & 32.94
& 38.99 & 83.79 \\

Dual GDN-MLA
& 509.80 & 39.95
& 35.17 & 125.05 \\

\midrule

Self-GDN + Cross-L-MLA
& 404.93 & 34.12
& 11.02 & 122.40 \\

Fused Self-GDN + Cross-L-MLA
& 375.84 & 32.18
& 11.04 & 110.82 \\

Triton Cross-GDN
& 478.57 & 40.19
& 34.26 & 126.38 \\

\bottomrule
\end{tabular}
}
\vspace{2pt}

\scriptsize
\emph{Note:} Dense Self-GDN: Self-$(G,G,G,\mathrm{SWA})\times3$ +
Cross-$(L,L,L,\mathrm{SWA})\times3$. Balanced GDN: Self-$(G,W,W,A_j)$ +
Cross-$(L,L,L,\mathrm{SWA})$ for $j=1,2,3$, with
$A_{1,2}=\mathrm{SWA}$ and $A_3=\mathrm{Full}$.
$G$: GDN; $W$: window attention; $L$: cached linear attention;
MLA: multi-head latent attention.

\end{table}

\section{Conclusion}
\label{sec:discussion}

Our results show that visual representation and history processing play
different roles in navigation world models. The preferred visual representation
changes with the prediction objective, showing that reconstruction quality
alone is not enough to choose the predictive state. History mixers, in
contrast, mainly differ in how they balance prediction quality and efficient
reuse of observation history. Global attention remains effective for simple
queries, while reusable history mechanisms are more suitable when the same
context is used across many candidate predictions. Overall, these findings
support a simple design principle: first choose the visual representation for
the prediction task, and then choose the history-processing method for the
expected workload.

\bibliographystyle{IEEEtran}
\bibliography{references}
\end{document}